\documentclass[10pt,twocolumn,letterpaper]{article}

\usepackage{authblk}

\usepackage[pagenumbers]{cvpr}
\usepackage{multirow}
\usepackage{fancyhdr}
\usepackage{array}
\usepackage{bbm}
\usepackage{graphicx}
\usepackage{amsmath}
\usepackage{amssymb}
\usepackage{booktabs}
\usepackage{bbding}
\usepackage{color}
\usepackage{appendix}
\usepackage{subcaption}
\usepackage{graphicx}
\usepackage[utf8]{inputenc}
\usepackage{CJKutf8}

\definecolor{cvprblue}{rgb}{0.21,0.49,0.74}
\usepackage[pagebackref,breaklinks,colorlinks,citecolor=cvprblue]{hyperref}
\usepackage{xcolor}
\definecolor{DarkGreen}{rgb}{0.43, 0.68, 0.28}

\newcommand\keywords[1]{\textbf{Keywords}: #1}

\title{NieR: Normal-Based Lighting Scene Rendering}

\author[1,2]{Hongsheng Wang\textsuperscript{*}}
\author[2]{Yang Wang\textsuperscript{*}}
\author[2]{Yalan Liu}
\author[2]{Fayuan Hu}
\author[1]{Shengyu Zhang \textsuperscript{$\dag$}}
\author[1]{\\ Fei Wu}
\author[2]{Feng Lin}

\affil[1]{Zhejiang University, China}
\affil[2]{Zhejiang Lab, China}

\makeatletter
\renewcommand\AB@affilsepx{, \protect\Affilfont}
\makeatother

\begin{document}
\begin{CJK}{UTF8}{gbsn}
\maketitle

\thispagestyle{fancy}

\lfoot{This work has been submitted to the IEEE for possible publication. Copyright may be transferred without notice, after which this version may no longer be accessible.}


\renewcommand{\headrulewidth}{0mm}

\renewcommand{\thefootnote}{\fnsymbol{footnote}}
\footnotetext[1]{These authors contributed equally to this work.}
\footnotetext[2]{Corresponding Author.}
\begin{abstract}

In real-world road scenes, diverse material properties lead to complex light reflection phenomena, making accurate color reproduction crucial for enhancing the realism and safety of simulated driving environments. However, existing methods often struggle to capture the full spectrum of lighting effects, particularly in dynamic scenarios where viewpoint changes induce significant material color variations. To address this challenge, we introduce NieR (\textbf{N}ormal-Based L\textbf{i}ghting Sc\textbf{e}ne \textbf{R}endering), a novel framework that takes into account the nuances of light reflection on diverse material surfaces, leading to more precise rendering. To simulate the lighting synthesis process, we present the LD (\textbf{L}ight \textbf{D}ecomposition) module, which captures the lighting reflection characteristics on surfaces. Furthermore, to address dynamic lighting scenes, we propose the HNGD (\textbf{H}ierarchical \textbf{N}ormal \textbf{G}radient \textbf{D}ensification) module to overcome the limitations of sparse Gaussian representation. Specifically, we dynamically adjust the Gaussian density based on normal gradients. Experimental evaluations demonstrate that our method outperforms state-of-the-art (SOTA) methods in terms of visual quality and exhibits significant advantages in performance indicators. Codes are available at \url{https://wanghongsheng01.github.io/NieR/}.

\end{abstract}


\section*{Introduction}
Recent advancements in 3D Gaussian Splatting technology (\textit{e.g.},~\cite{aso3d2024}) have led to significant improvements in visual quality and rendering efficiency by employing pixel-level Gaussian distributions for scene rendering.  While existing methods demonstrate excellence in numerous areas, they still struggle with dynamic environmental lighting changes~\cite{Interactive}, particularly when rendering challenging specular objects such as vehicles in autonomous driving scenarios. This often leads to distortion and blurriness, underscoring the ongoing challenge of achieving real-time and precise rendering of lifelike images in this field.

In nature, when lighting hits on the surface, both specular reflection(refs. ~\cite{specular2004,tfor1967}) and diffuse reflection (refs. ~\cite{Idiffuse1998}) occur simultaneously. The core of 3D Gaussian rendering relies on using a continuous, anisotropic Gaussian function to process each point in an image(refs. ~\cite{Deblurring,Deferred}). However, specular reflection is a precise process, where the direction of reflected light strictly depends on the angle at which the light hits the surface and the angle at which we observe it. The uniform smoothness of the Gaussian distribution makes it hard to accurately capture and represent this complex interaction of lighting. Additionally, specular surfaces typically exhibit high brightness contrast. Traditional 3D Gaussian rendering techniques struggle to effectively handle such high dynamic range scenes, resulting in overexposure in highlight areas or lack of details. In real-world environments, diffuse reflection is influenced not only by direct illumination but also by various factors such as ambient light and indirect light (\textit{e.g.}, reflected light). For diffuse reflection, accurately simulating the complex interactions between light sources and object surfaces is crucial, while 3D Gaussian rendering often focuses on enhancing visual effects through pixel-level Gaussian Splatting, failing to capture the complexity of scattered lighting.

The Normal-Based Lighting Scene Rendering model introduces an optimized Physically Based Rendering (PBR)  method through lighting decomposition to enhance the color synthesis and densification strategy of 3D Gaussian, which was able to improve the rendering quality and realism of 3D Gaussian for specular objects. Firstly, we propose Light Decomposition 3D Gaussian (LD) to optimize the color synthesis method by decomposing the outgoing radiation into specular reflection and diffuse reflection by means of surface normals. We also introduce the specular reflection coefficient attribute to consider the degree of lighting reflection by materials, significantly enhancing the accuracy of rendering specular objects and the realism of lighting effects. However, due to the sparse Gaussian representation of scenes in 3D Gaussian, significant large displacement changes will occur during optimization. Describing bright details poses difficulty in capturing ample lighting information, particularly for highlights, making it challenging to ensure the overall quality of rendered images. To address this deficiency, we introduce the Hierarchical Normal Gradient Densification (HNGD). NieR can adaptively adjust the density of Gaussian points based on the changing gradients in the scene further improving the accuracy of rendering specular reflective objects and effectively addressing the limitations of traditional sparse Gaussian representations in capturing highlights details.\\
In summary, our main contributions are as follows:\\
$\bullet$ We proposed a lighting decomposition module LD to optimize the color synthesis method of 3D Gaussian.\\
$\bullet$ We introduced hierarchical normal gradient densification to address the issue of insufficient lighting information caused by sparse Gaussian points, further improving the rendering quality of specular objects.\\
$\bullet$ Extensive experiments validate that the combination of LD and HNGD methods can significantly enhance the details of scene rendering. Especially in handling specular reflections and complex lighting scenarios, our method achieves the best visual quality compared with those of the state-of-the-art (SOTA) methods.

\section{Related Work}
\subsection{Traditional Physical Rendering Methods} 

Traditional physical reconstruction techniques~\cite{srm2004,Imphong,blinnphong,PAblinnphong} strive to achieve highly realistic image rendering by finely modeling the complex physics of the interaction between material surfaces and light. In the initial stages, techniques such as the Lambertian~\cite{lambert} reflection model are based on the assumption of ideal diffuse reflection surfaces to achieve basic simulation of lighting effects. However, this model is inadequate for representing more complex phenomena such as highlights and anisotropic reflection. Therefore, the field introduced the Phong~\cite{phong} reflection model and its improved version, the Blinn-Phong model, which approximate lighting effects using empirical formulas and consider both reflected light and material properties, significantly enhancing the ability to simulate lighting phenomena.

A pivotal role in this domain is held by the Bidirectional Reflectance Distribution Function~\cite{Evmt2017,MMob2013} (BRDF), quantifying the scattering or reflection behavior of specific incident and reflected light directions, thereby offering a theoretical foundation for simulating diverse surface characteristics. The Cook-Torrance model [citation needed], as an application, combines microfacet theory and the Fresnel effect to provide accurate lighting simulations for rough surfaces, while the Disney Principled BRDF [citation needed] simulates a wide range of materials with controllable parameters (\textit{e.g.}, base color, metallicity, and roughness), which further enhances fidelity and richness of detail in image rendering. 

\subsection{implicit Rendering Methods}
In contrast to traditional techniques, Neural Radiance Fields (NeRF)~\cite{NerF} leverage implicit Multilayer Perceptrons (MLPs) to render density and view-dependent colors from three-dimensional coordinates and viewing directions~\cite{miner,nerfds,Nerfies}, advancing the development of differentiable volume rendering. In an extended study, a method is introduced for reconstructing high-quality geometric shapes and complex spatially varying BRDFs from a sparse image datasets, which integrating and refining geometry and BRDF by optimizing the latent space of a reflectance network, significantly reducing photometric differences. This approach also demonstrates practicality on sparse image datasets and significant progress in acquiring high-quality shapes and appearances. While NeRF exhibits unprecedented capabilities in reconstructing and rendering high-quality 3D scenes, it still faces the challenge of high computational costs in practical applications.~\cite{Anerf,VMNeRF,Adop,HDR-Plenoxels}

\subsection{3D Gaussian-based Methods}
3D Gaussian Splatting~\cite{3DGSrealt} directly constructs 3D scenes by employing a three-dimensional Gaussian distribution, which significantly surpasses NeRF in rendering speed, particularly in the optimization process of light synthesis. Relightable~\cite{Relightable3dgs} 3D Gaussian extends its functionality by combining 3D Gaussian points with normals, global and local quantities of incident light, and BRDF parameters, refined through physics-based differentiable rendering. This method also introduces an efficient visibility baking technique based on boundary volume hierarchies~\cite{boundary,Bounding},  enabling real-time rendering and precise shadow effects. GaussianShader~\cite{GaussianShader} introduces a simplified shading function tailored for 3D Gaussians, focusing on enhancing the rendering effects with scenes containing reflective surfaces while ensuring the efficiency of both training and rendering processes. By introducing an innovative normal estimation framework, this method tightly associates shading parameters with the true geometric form of 3D Gaussians, significantly improving the rendering quality, especially when rendering specular objects. Nevertheless, challenges still exist in accurately simulating light reflection effects, particularly in precisely capturing specular reflective objects, occasionally leading to visual distortion and deformation of reflective objects in reconstructed scenes. Our method takes into account the differential effects brought about by different reflections in the scene and separately handles the reflective effects in the scene.

\section{Methodology}

\subsection{Preliminary}
\textbf{3D Gaussian Splatting. } Unlike traditional NeRF (Neural Radiance Fields) implicit neural  networks, 3D Gaussian employs several three-dimensional Gaussian  distributions to explicitly represent scenes, thereby effectively  improving the rendering speed. In mathematics, a three-dimensional  Gaussian distribution can be defined by the following formula:\\
\begin{equation}
G(x)=exp(-\frac{1}{2}(x-μ)^T\textstyle \sum^{-1} (x-μ))
\end{equation}

Among them, $\mu$ is the center position, and $\Sigma$ is the 3D covariance matrix. In addition, 3D Gaussian Splitting incorporates information on opacity $o$ and color $c$ into each Gaussian distribution for subsequent rendering work. In our data, color $c$ is represented by a fourth-order spherical harmonic function~\cite{Sharmonics,Spherical}.
Subsequently, we project the three-dimensional Gaussian onto a two-dimensional plane to render the image.\\
\begin{equation}
\textstyle \sum' = JW\Sigma W^TJ^T
\end{equation}

Among them, $\Sigma'$ is the 2D covariance matrix in the camera coordinates, $J$ is the Jacobian matrix~\cite{jacobian} of the affine approximation projection transformation, and $W$ is the world-to-camera perspective transformation matrix.
Given the position of pixel $x$, we solve for its final color by mixing N ordered points that overlap with the pixel. 
\begin{equation}
c=\sum_{i∈N}T_i \alpha_ic_i \quad with\quad T_i=\prod_{j=1}^{i-1}(1-\alpha_j)
\end{equation}

Here, $\alpha$ is obtained by multiplying the contribution of opacity $o$ with the 2D covariance calculated by $\Sigma'$ and the image's spatial pixel coordinates.

\textbf{Rendering equation} The rendering equation~\cite{rendering} is used to describe the interaction between light  and surfaces and is widely used in lighting models in the physical  world. It calculates the radiance at surface points $x$ along the observation direction:
\begin{equation}
L_o(\omega _o,x)=\int_{\Omega }^{} f(\omega _o,\omega _i,x)L_i(\omega _i,x)(\omega _i\cdot n)d\omega _i
\end{equation}

Here, $x$ is the surface point, $n$ is the surface normal vector,  $f$ is the BRDF (Bidirectional Reflectance Distribution Function) that  describes the spatial reflection characteristics of the surface, and  $L_i$ is the ray incident from direction $\omega_i$ onto point $x$, integrated over the hemisphere in direction $\Omega$ towards  $\omega _i$. When rendering 3D scenes, it is important to choose the  appropriate method for $f$ and $L_i$ due to the complex properties  of materials and the differences in lighting in different scenes.

\begin{figure*}[htbp]
    \centering
    \includegraphics[width=\textwidth]{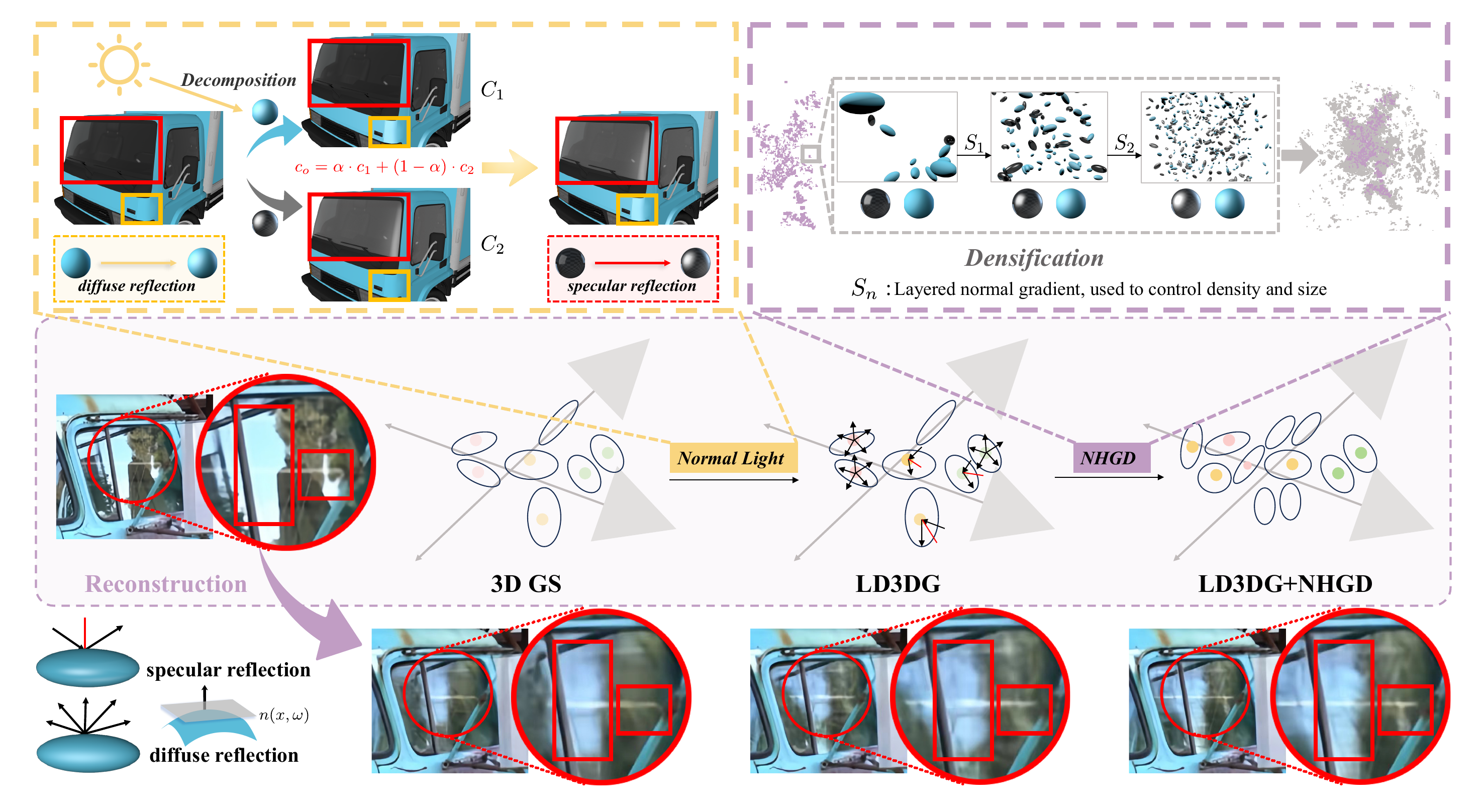}
    \caption{The pipeline of our method, with the LD module on the left and the HNGD module on the right}
    \label{fig:pipeline}

\end{figure*}

\subsection{LD: Light Decomposition 3D Gaussian }

In the field of three-dimensional scene rendering, achieving highly realistic visual effects not only mandates accurate simulation of lighting effects but also demands a thorough consideration of the differences between diffuse~\cite{Idiffuse1998} reflection and specular~\cite{specular2004} reflection lighting phenomena. This requirement is particularly prominent when dealing with materials that exhibit different lighting response characteristics, such as rough wooden surfaces and smooth metallic surfaces. However, current methods often overlook these details, leading to significant discrepancies between the rendered results on complex material surfaces and the real world.

Therefore, when dealing with complex lighting scenarios, the challenge lies in how to accurately capture the different effects of diffuse reflection and specular reflection while ensuring computational efficiency, and take full account of the interaction between the surfaces, changes in lighting conditions and the diversity of material properties~\cite{Gms2018}. In this light, we propose an efficient three-dimensional scene lighting framework that integrates diffuse reflection, specular reflection, and incident light models. In this model, we employ a simplified incident lighting model $c$ to represent the impact of light on the scene.
\begin{equation}
c = V \cdot L_{global} + L_{local}
\end{equation}

To enhance the realism and naturalness of the rendering effects, we have refined the simulation of diffuse reflection and specular reflection. Firstly, by combining spherical harmonics and the cosine of the angle between the light ray and the surface normal, we carefully adjust the contribution of diffuse reflection to the lighting effects to ensure improved naturalness and realism of the lighting effects from different viewing angles. Additionally, we have further refined the application of spherical harmonics in predicting specular reflection by more precisely controlling the direction of lighting reflection and considering the influence of surface properties on reflected lighting, that leads to a more realistic specular highlights. Furthermore, considering the differences in how different materials react to lighting, we have carefully adjusted the balance between diffuse reflection and specular reflection in the rendering process. Specifically, we utilize $c_0$ to calculate the final rendered color.

\begin{equation}
c_0=(1-a)\cdot shc \cdot cos\theta + a\cdot shc
\end{equation}

Where $a$ is the weighting factor used to adjust the ratio of the two reflection effects in the final result, when $a = 1$ means that favor towards full specular reflection and $a = 0$ indicates the pure diffuse reflection; $shc$ is the spherical harmonic coefficient used to calculate the colors for diffuse and specular reflections; $\cos(\theta)$ takes into account the angle between the direction of the lighting and the surface normal, affecting the intensity of diffuse reflection.

\subsection{HNGD : Hierarchical Normal Gradient Densification}

 The introduction of the spherical harmonic function allows us to quickly estimate the light intensity in any direction from precomputed coefficients, providing an effective means of approximating the lighting in complex scenes. However, when these Gaussian points are too sparse, the light prediction may produce large errors, especially in specular reflections and highlight details, due to the lack of sufficient local light information. To reduce this error and optimize the lighting prediction, we introduce HNGD (Hierarchical Normal Gradient Densification) for finer lighting control to achieve higher quality rendering.
 
 \text{Hierarchical Gradient Densification.} Under empirical study, we found that simply increasing the number of Gaussian points does not necessarily improving result accuracy. On the contrary, this approach may lead to unnecessary rendering errors due to the mutual interference between Gaussian sampling distributions. In order to solve this problem, we propose a hierarchical gradient densification approach, which allows us to design different sizes of gradient deflators for specific needs and features of the scene to further improve the effectiveness of the densification process.

Specifically, during the experimental process, when the fused gradient values of a certain region within the Gaussian processed by LD reach the threshold for the next level, we proceed with a splitting operation on the relevant Gaussian. This operation involves dividing each Gaussian into two, aiming to refine the distribution of Gaussian. By controlling the number of Gaussian in a manner most beneficial for the current region's lighting representation, this approach allows for a more precise capture of lighting details in complex scenes.

\textbf{Normal Gradient Fusing.} We adjust the densification of the Gaussian points by using the gradient information. However, since 3D Gaussian are more sparsely distributed in space compared to traditional point cloud data, the geometric gradient of the scene does not realistically represent the shape changes. For this reason, we perform a weighted fusion of geometric and normal vector gradients and use adjustable weighting coefficients to meet different experimental needs. 

First, we use the center difference method for each volume point (x,y,z) in the scene to compute its gradient components in the three directions X, Y, and Z.

\begin{equation}
\begin{aligned}
 G_x & = V(x+1, y, z) - V(x-1, y, z)  \\
  G_y & = V(x, y+1, z) - V(x, y-1, z) \\
   G_z & = V(x, y, z+1) - V(x, y, z-1)  \\
\end{aligned}
\end{equation}

where V(x, y, z) is the density value located at (x, y, z). Next, we normalize the gradient vector $G_{xyz} = (G_x, G_y, G_z)$ to obtain the unit normal vector $ \hat{G_{xyz}} = \frac{G_{xyz}}{\|G_{xyz}\|} $,$ \|G_{xyz}\|=\sqrt{G_x^2 + G_y^2 + G_ z^2} $.
Further, we compute the normal vector gradient $G_{norm}$ to capture the gradient information induced by the change of normal vector:

\begin{equation}
\begin{aligned}
 \Delta \hat{N}_x & = |\hat{N}(x+1, y, z) - \hat{N}(x, y, z)|  \\
  \Delta \hat{N}_y & = |\hat{N}(x, y+1, z) - \hat{N}(x, y, z)| \\
  \Delta \hat{N}_z & = |\hat{N}(x, y, z+1) - \hat{N}(x, y, z)|  \\
\end{aligned}
\end{equation}

Combining the computed normal vector gradient components $\Delta \hat{N}_x$, $\Delta \hat{N}_y$, and $\Delta \hat{N}_z$, we obtain the normal vector gradient vector $\Delta \hat{N} = (\Delta \hat{N}_x, \Delta \hat{N} _y, \Delta \hat{N}_z)$. The modulus of $\Delta \hat{N}$ is taken to obtain a scalar representation of the normal vector gradient strength: $G_{norm} = \| \Delta \hat{N} \|$.
We use the following formula for weighted fusion:
\begin{equation}
Grad = (1 - \omega) \cdot \left( \frac{G_{xyz}}{denom} \right) + \omega \cdot \left( \frac{G_{norm}}{denom} \right)
\end{equation}
where denom is the normalization factor and $\omega$ is the fusion weight, between 0 and 1, used to regulate the relative importance of the scene gradient (geometric gradient) (accumulation of) $G_{xyz}$ and the normal vector gradient (accumulation of) $G_{norm}$ in the final fusion gradient $Grad$.

\section{Experimentation}
\subsection{Training Detail}

Many advanced methods exist for reconstructing and rendering real-world scenes (\textit{e.g.}, ~\cite{4dgaussian, Dynamic3dgs,nerfds,Human101}).  We discuss related work in detail in Section 2. We implemented our method using PyTorch and CUDA, leveraging a single NVIDIA V100 GPU. Following the training protocol of 3D Gaussian Splatting (3D GS) ~\cite{4dgaussian}, all models were trained for 30,000 iterations, with each scene reconstruction requiring less than 3 hours. We extend 3D GS by introducing additional parameters to control the synthesis of diffuse and specular reflections, enabling fine-grained control over lighting components during the rendering process.

\begin{figure*}[htbp]
    \centering
    \includegraphics[width=\textwidth]{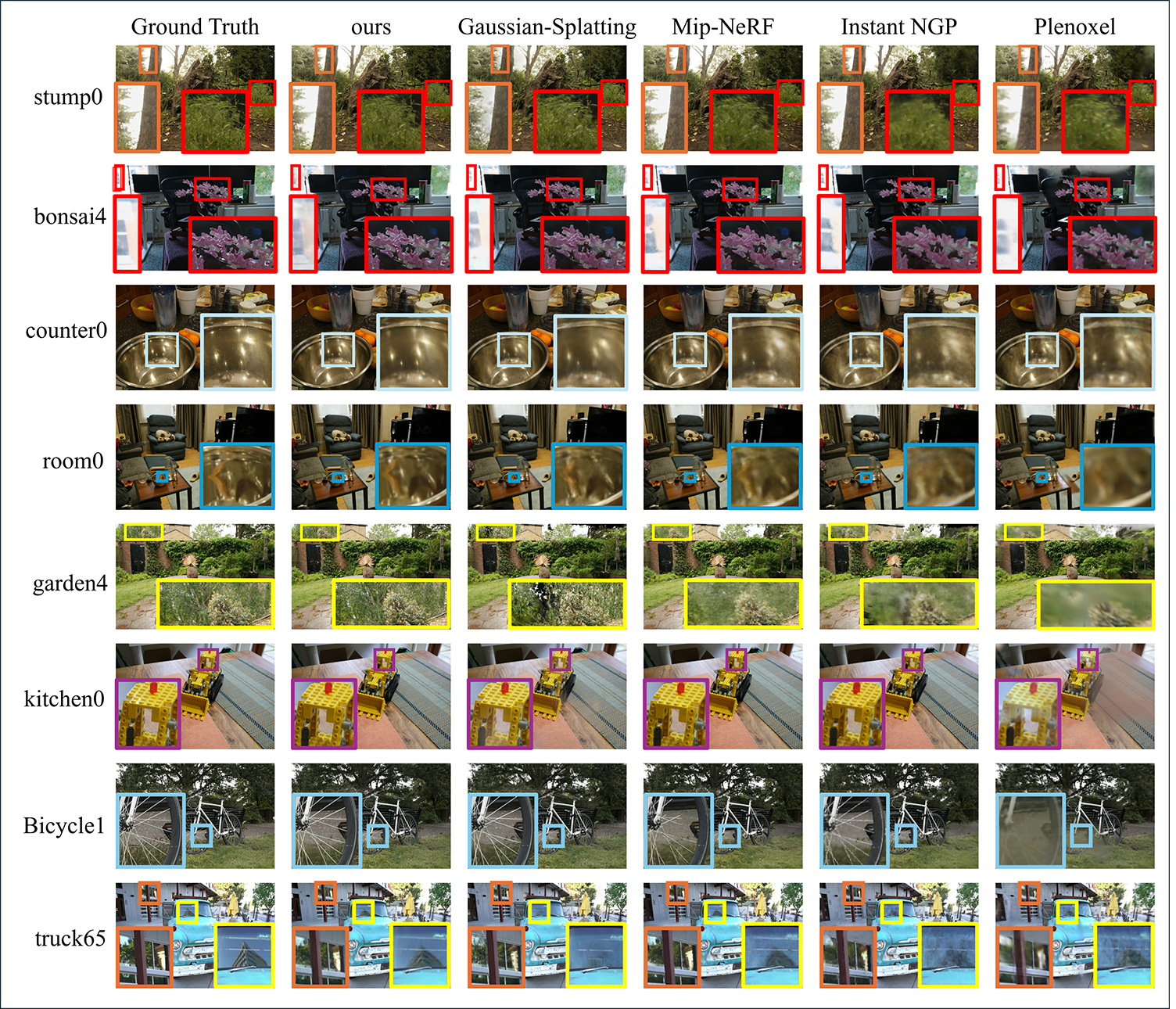}
    \caption{Shows the comparison between our method and previous methods, as well as the corresponding ground truth images extracted from the preserved test views. The scenes from top to bottom are the stump, bonsaimcounter, room, garden, kitchen, and cycle from the Mip NeRF360 dataset; And trucks and trains from Tanks and Temples. Non obvious quality differences are marked with arrows/embedded markings.}
    \label{fig:cmp}

\end{figure*}
\subsection{Results and Evaluation Results}

To assess NieR's ability to render scenes with realistic lighting and fine details, we compare it against Gaussian Splatting as the baseline method.  We employ established quantitative metrics, including Peak Signal-to-Noise Ratio (PSNR) ~\cite{PSNR}, Learned Perceptual Image Patch Similarity (LPIPS) ~\cite{LPIPS}, and Structural Similarity Index (SSIM) ~\cite{SSIM}. These metrics evaluate the similarity between rendered images and ground truth data, quantifying visual fidelity and detail preservation. we compare our method with Gaussian Splatting, Mip-NeRF360~\cite{mipnerf360}, InstantNGP, and Plenoxels on seven scenes from the Mip-Nerf360 dataset~\cite{mipnerf360} and two scenes from the Tanks$\And$Temples dataset [2017]. This diverse set of scenes includes both indoor environments with stable lighting (living rooms, kitchens) and outdoor scenarios characterized by complex lighting conditions (playgrounds, gardens).  Quantitative comparisons are presented in Table \ref{tab:1} and Table \ref{tab:2}, while Figure \ref{fig:cmp} showcases qualitative comparisons. NieR demonstrates competitive performance against baselines across various scenes, particularly in scenarios characterized by \textbf{complex lighting}. Notably, our method achieves substantial improvements in both quantitative metrics and visual quality under such challenging conditions, highlighting its effectiveness in rendering scenes with \textbf{realistic lighting and intricate details}.

\begin{table*}[tb]
    \centering
    \caption{The average metrics.We report the results of our model and other
     models compared on the Mip NeRF360 dataset and Tanks $\And$ Temples dataset in all scenarios. The best number is highlighted in bold. ↑ indicates that the larger the indicator, the better, ↓ indicates that the smaller the indicator, the better.}
    \label{tab:main1}

    \resizebox{\textwidth}{!}{%
    \begin{tabular}{l|c|c|c|c|c|c|c|c}
    \toprule
    \multirow{2}{*}{Model} & \multicolumn{4}{c}{Mip NeRF360} & \multicolumn{4}{|c}{Tanks $\And$ Temples} \\
    & SSIM↑ & PSNR(dB)↑ & LPIPS↓ &Train & SSIM↑ & PSNR(dB)↑ & LPIPS↓ &Train \\
    \midrule
    Plenoxels & 0.686& 23.067 & 0.386& 25m49s & 0.7180	& 21.08	& 0.343& 25m5s\\
    InstantNGP & 0.724 & 25.381 & 0.336& 5m37s &  0.7225	& 21.71	& 0.329& 5m26s\\
    Mip-NeRF & 0.823 & 27.585 & 0.219 & 48h &0.7575	&22.21	&0.256&48h\\
    Gaussian-Splatting & 0.863 & 27.455 & 0.181 & 41m33s & 0.8400	& 23.14	& 0.182 & 26m54s\\
    ours &\pmb{0.910} &\pmb{30.524} &\pmb{0.153} & 1h36m & \pmb{0.9170}	&\pmb{28.18}	&\pmb{0.132}&1h59m\\
    \bottomrule
    \end{tabular}
    }
\label{tab:1}
\end{table*}

\textbf{Comparisons with other Methods. }As one of the most competitive NeRF methods, the Mip-NeRF360 method estimates color and brightness by integrating across anti-aliased cone regions of lighting. However, it does not consider the reflective properties of different materials, which leads to suboptimal performance in scenes with specular objects. Because of the 3D Gauss Splatting method represents the scene using 3D Gaussians and performs interleaved optimization and density control on these Gaussians, providing an accurate representation of the scene. Nevertheless, its reliance solely on spherical harmonics for lighting processing results in coarse rendering effects in complex lighting environments. Our model introduces a lighting synthesis module and a hierarchical densification module, offering a more effective rendering method for complex lighting scenes and maintaining strong competitiveness in scenes with relatively stable lighting. As shown in Table \ref{tab:1}, our method achieves the highest level across all metrics. Compared to Mip-NeRF, our method improves SSIM by 8.7 percent, and compared to 3D Gauss Splatting, it improves SSIM by 4.7 percent. The result indicates that our method not only effectively reproduces realistic lighting variations but also reduces the interference of noise on image quality, and realistic the restoration of picture colors.

\begin{table*}[tb]
    \centering
    \caption{Our model and other comparison models have PSNR metrics for all scenarios in the Mip NeRF360 dataset and Tanks/Temples dataset.}
    \label{tab:main2}

    \resizebox{\textwidth}{!}{%
    \begin{tabular}{l|l|l|l|l|l|l|l|l|l|l}
    \toprule
    dataset & Stump & Room & Kitchen & Garden & Counter & Bonsai & Bicycle & truck & train & average \\
    \midrule
    Plenoxels           &20.677             &27.574             &23.433             &23.503             &23.636             &24.710             &21.897  &23.231 &18.944          &23.067 \\
    InstantNGP          &23.625             &29.269             &28.547             &25.068             &26.438             &30.337             &22.192  &23.260&20.170          &25.434\\
    Mip-NeRF360         &26.356             &\pmb{31.467}             & 31.988             &26.875             & 29.447             &\pmb{33.396}             &24.304   &24.911&19.522          & 27.585 \\
    Gaussian Splatting  & 26.545             &30.632             &30.316             & 27.410             &28.692             &31.980             & 25.245             & 25.186 & 21.097& 27.455\\
    ours                &\pmb{30.984}             &\ 30.886             &\pmb{33.484}             &\pmb{29.661}             &\pmb{32.573}             & 33.369             &\pmb{27.398}  &\pmb{29.993} &\pmb{26.376}           &\pmb{30.524} \\
    \bottomrule
    \end{tabular}
    }
\label{tab:2}
\end{table*}

To further explore the performance of our method in different scenes, we present the comparative results based on the PSNR metric in Table \ref{tab:2}. It can be observed that our method shows significant improvements in outdoor scenes like gardens and in scenes containing specular objects such as trains and cars. Especially in the train scene with a large number of specular objects, our method outperforms Gaussian Splatting by 25.02 percent in terms of PSNR. This indicates a substantial enhancement in rendering specular object scenes with our proposed method. Additionally, our method surpasses our baseline method Gaussian Splatting in all scenes. While Mip-NeRF can produce high-quality results in some scenes, it requires an average training time of 48 hours on the same device. In contrast, our method has an average training time of 1 hour and 47 minutes. It shows that our method not only achieves high-quality rendering effects but also maintains \textbf{low memory consumption and rendering time}.

\textbf{Qualitative Results. }InstantNGP~\cite{Instantngp}, Plenoxels~\cite{Plenoxels},3D Gaussian~\cite{3DGSrealt}, and Mip-NeRF360~\cite{mipnerf360} rely solely on the viewing direction for rendering, resulting in the loss of some lighting information, which will cause the potential boundary blurring and artifacts on the rendered object surfaces, as exemplified in the truck scene. While frameworks such as Mip-NeRF360 can render the outline of a building on the truck's glass, it fails to capture details regarding the floors and boundaries of the building. To address this deficiency of lost lighting information, our framework introduces specular and diffuse reflection information present on the object surface. Our comparison shows that we can more accurately reproduce the lighting changes and the clear outlines of building on the truck's glass. 3D Gaussian utilizes Gaussian distribution for pixel smoothing during scene 

rendering, leading to halo-like diffusion on surfaces under complex lighting conditions. For example, the "counter" scene in Figure \ref{fig:cmp}, which smooth glare spots appear on the inner wall of the basin, lacking the expected gloss and brightness in the central points. In order to mitigate the impact of Gaussian smoothing, our framework employs densification of Gaussian in areas with stronger lighting, resulting in brighter light spots on the basin surface. In summary, these scenes demonstrate the advantages of our framework in reproducing realistic lighting environments. For instance, when generate color characteristics of materials such as metal and glass under lighting, our framework provides surfaces with more gloss and brightness, approaching closer to the ground truth.

\subsection{Ablation Studies}

We segregate the algorithm into distinct modules and further decomposed each module. Simultaneously, we design a set of experiments to investigate the significance of each module and its corresponding submodules on the model. Ablation studies will be conducted on the LD and HNGD modules. 

\subsubsection{Light Decomposition}

\begin{figure}[htbp]
    \centering
    \includegraphics[width=0.5\textwidth]{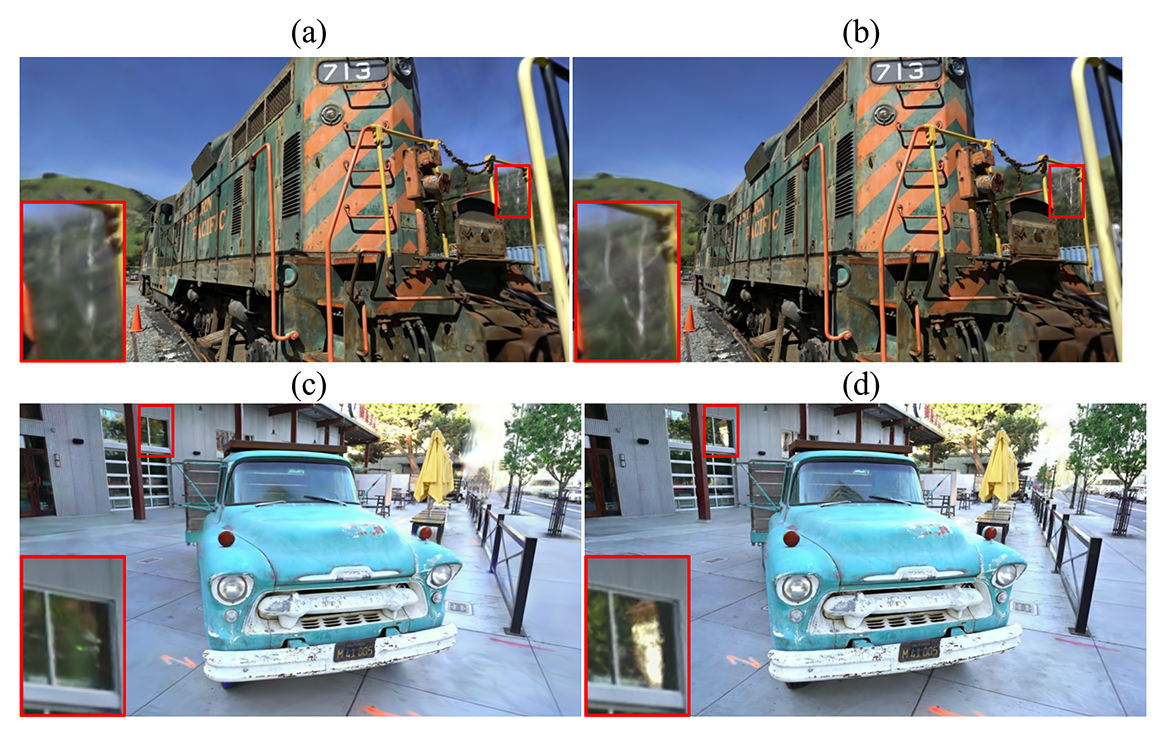}
    \caption{The top left image (a) shows the result of removing diffuse light intensity, while the top right image (b) shows the complete result. The left image (c) shows the result of removing specular reflection, while the right image (d) shows the complete result.}
    \label{fig:3}
\end{figure}

\paragraph{\textbf{Comparisons with Baseline}}It would fail to accurately reconstruct details in \textbf{scenarios involving diffuse or specular reflections}, due to the limitation of 3D Gaussian Splatting, which relies solely on spherical harmonics for handling surface lighting.To validate the ability of the lighting decomposition module to recreate different lighting conditions in a scene, we conducted experiments on the train and truck scenarios, as shown in Figure \ref{fig:3} (left column shows the rendering results of 3D Gaussian Splatting, while the right column displays our results). Firstly, Compared with images (a) and (b), our framework eliminates the shadow artifacts cast by deep green trees on the white steel frame, revealing clearer outlines of the steel structure. Secondly, for the windows of the house shown in images (c) and (d), representing reflections on specular material objects under complex lighting conditions, our module not only reconstructs the tree information with real light and shadow changes in the left window but also renders bright lighting information distinct from other objects in the right window. Finally, after multiple comparisons of rendered images from different perspectives, it was observed that our framework can \textbf{faithfully reproduce the color characteristics} of object surfaces, whether in scenarios involving diffuse reflections or more pronounced specular reflections, which proves the effectiveness of integrating diffuse reflection, specular reflection, and incident light models as input information.

\paragraph{\textbf{Quantitative Results}}To quantitatively evaluate the rendering capability of the LD module in addressing dynamic lighting perception, we conducted an ablation experiment on the original scheme by removing the LD module to observe changes in metrics. As shown in Table \ref{tab:3}, \textbf {after the LD module is removed, the PSNR and SSIM metrics decreased by 5.5 percent and 12.8 percent , while the LPIPS metric increased by 17.6 percent.} The significant changes in LPIPS and SSIM metrics resulting from the removal of the LD module, indicate that the introduction of lighting information can produce images closer to the ground truth in terms of human visual perception, and accurately simulating the color characteristics of object surfaces under complex lighting conditions.

\begin{table}[h]
    \centering
    \caption{ Ablation studies on NeRF360 datasets using our proposed methods.}
    \label{tab:ablation1}
    \resizebox{0.5\textwidth}{!}{%
    \begin{tabular}{l|l|l|l}
    \toprule
    Model & PSNR(dB)↑ &SSIM↑ &LPIPS↓\\
    \midrule
    Ours w/o LD and HNGD   &0.878&25.186&0.147 \\
    Ours w/o HNGD          &0.929&28.886&0.125\\
    ours                   &\pmb{0.945} &\pmb{30.19} &\pmb{0.090} \\
    \bottomrule
    \end{tabular}
    }
    \label{tab:3}
\end{table}

\paragraph{\textbf{Analysis on the  Lights}}To validate the capability of the LD module in simulating real lighting scenarios, we compared the scene without the diffuse reflection effect and without the specular reflection effect by adjusting the parameters in the LD module. It was observed that when removing either the diffuse reflection effect or the specular reflection effect, the metrics of the model will decline. This further emphasizes the effectiveness and importance of the LD module in simulating lighting effects(especially specular reflection). Additionally, by comparing the changes in the LPIPS metric during the ablation process, we found that the complete LD module better preserves the contours and colors of buildings in specular reflections, making the rendered images closer to the original ground truth.

\subsubsection{Hierarchical Densification }

\begin{figure}[h]

    \centering
    \includegraphics[width=0.5\textwidth]{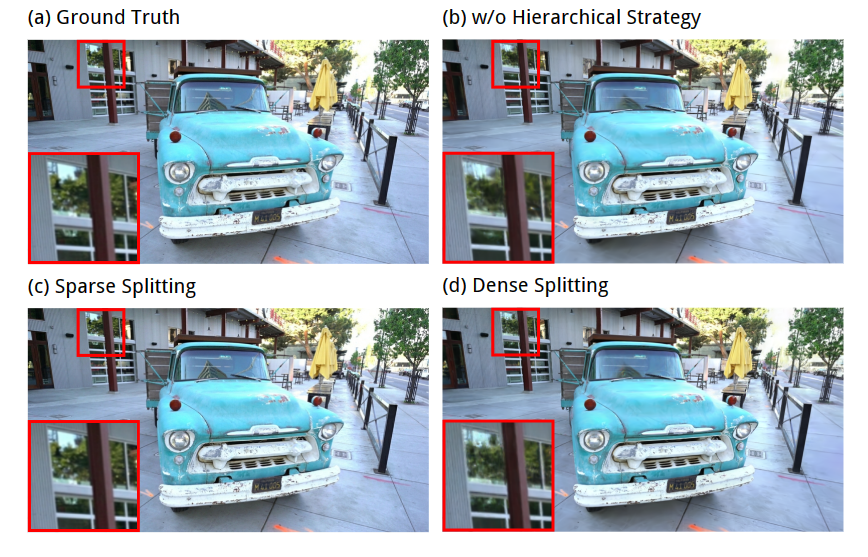}

    \caption{From left to right (a) shows Ground Truth, (b) shows no hierarchical densification result, (c) shows sparse splitting hierarchical densification result, and (d) shows dense splitting hierarchical densification result.}
    \label{fig:ablation_ld}

\end{figure}

\paragraph{\textbf{Effects of HNGD module}}Furthermore, we explored the role of the hierarchical densification module in addressing lighting variations. In Figure \ref{fig:ablation_ld}, we present visual comparison results of three methods: without hierarchical densification strategy (b), sparse hierarchical densification strategy (c), and dense hierarchical densification strategy (d). In the experiment, we introduced a predefined threshold range [1, 1.5, 2, 2.5, 3, 3.5] as the discriminative thresholds for the sparse splitting hierarchical strategy, with values greater than 3.5 being considered for the dense splitting hierarchical strategy. It was observed that the method without the addition of the hierarchical densification module struggled to render high-quality results, as seen in image(b). Compared with the ground truth revealed issues such as the fogging of the station skeleton in the upper left corner and the trees to the right, significant blurriness at the boundaries of mapped trees, subtle variations in brightness, unclear color patches in the darker leaf area at the bottom right of the mirror, and a loss of hierarchical relationships with shadows.
\begin{table}[h]
    \centering
    \caption{Ablation studies on the impact of lighting}
    \label{tab:ablation2}
    \resizebox{0.5\textwidth}{!}{%
    \begin{tabular}{l|l|l|l}
    \toprule
    Model & PSNR(dB)↑ &SSIM↑ &LPIPS↓\\
    \midrule
    Light Synthesis w/o Diffuse Reflection   &0.928	&28.791	&0.129 \\
    Light Synthesis w/o specular &0.905	&26.771	&0.161\\
    Full Light Synthesis &\pmb{0.929}	&\pmb{28.88}	&\pmb{0.125} \\
    \bottomrule
    \end{tabular}
    }
\end{table}

Our framework effectively  addresses this issue by employing the hierarchical densification module, which can selectively generate more Gaussian in regions with strong brightness variations. Compared with Figure \ref{fig:ablation_ld}(a), (b), and (c) reveals that even with the use of sparse hierarchical densification strategy, we can accurately render a station with a brighter skeleton and surrounding leaves, as well as bring out the tree leaves in the lower right area of the mirror from the dark shadow. Additionally, compared with Figure \ref{fig:ablation_ld}(a), (c), and (d), we observe that as the densification level of Gaussian points increases, the trees on the windows exhibit clearer brightness variations against a bright sky background, and capture more light spots projected between the leaves. Analysis of the above comparative experimental results, we have a significant observation that lighting scenes without using our HNGD module, there is often a phenomenon of objects blending together and resulting in color fog. However, when we adjust the Gaussian density using HNGD, we render scenes with realistic hierarchical details, demonstrating the effectiveness of our method in achieving realism in rendering.

To quantitatively evaluate the impact of the hierarchical densification module on enhancing the rendering capability for dynamic lighting perception, we demonstrated the gains brought by this module in terms of metrics by removing it from the existing method. As shown in Table \ref{fig:5}, our method shows improvements in multiple metrics compared to the method without the densification module. Specifically, the method using dense hierarchical strategy experiences a 30.9 percent decrease in LPIPS metric, demonstrate the restoration capability of this module in lighting scenes.

\begin{table}[h]
    \centering
    \caption{Ablation studies on the effect of hierarchical strategy  to densification}
    \label{tab:ablation2}
    \resizebox{0.5\textwidth}{!}{%
    \begin{tabular}{l|l|l|l}
    \toprule
    Model & PSNR(dB)↑ &SSIM↑ &LPIPS↓\\
    \midrule
    ours w/o  hierarchical strategy   &0.928	&28.791	&0.129 \\
    ours with sparse splitting hierarchical strategy &0.938	&29.592	&0.106\\
    ours with dense splitting hierarchical strategy &\pmb{0.945}	&\pmb{30.195}	&\pmb{0.090}\\
    \bottomrule
    \end{tabular}
    }
\label{fig:5}
\end{table}

\paragraph{\textbf{Effects of Different Hierarchical Strategies}}To validate the influence of our hierarchical densification module on perceiving different lighting variations through hierarchical adjustments, we dynamically adjust the densification levels in the lighting variation module to observe the effects of different settings in handling various lighting scenarios. The image in the red box in Figure \ref{fig:ablation_ld} shows that, although dense densification has a better build-up of the structure of reflective objects such as the base station, there is no significant change for the deep shadow parts of the tree leaves compared to sparse densification. This result guides us to adaptively adjust the densification strategy based on the strength of lighting in the scene to save unnecessary time and memory overhead. In conclusion, as the level of densification perceived by the module increases in areas with significant lighting, it notably enhances the presence of stronger brightness variations in corresponding positions in the rendered images.


\section{Conclusion}
 This paper presents NieR, a novel rendering method that incorporates both diffuse and specular reflection information. NieR addresses regions with significant illumination variations through a hierarchical densification strategy, leading to demonstrably improved rendering accuracy and realism. This is particularly evident for objects with specular properties and complex scenes. Our work paves the way for incorporating lighting information into 3D scenes rendered using the Gaussian field representation. Future work will focus on exploring NieR's applicability in dynamic lighting scenarios and investigating its potential integration with other rendering techniques for further performance gains.

\newpage
\bibliographystyle{ieeenat_fullname}
\bibliography{main}

\end{CJK}
\end{document}